\documentclass{article}

\usepackage{PRIMEarxiv}

\usepackage[numbers,sort]{natbib}

\usepackage[utf8]{inputenc} 
\usepackage[T1]{fontenc} 
\usepackage{hyperref} 
\usepackage{url} 
\usepackage{booktabs} 
\usepackage{amsfonts} 
\usepackage{nicefrac} 
\usepackage{microtype} 
\usepackage{multirow}
\usepackage{graphicx}
\usepackage{float}
\usepackage{subcaption}
\usepackage{algorithm}
\usepackage{algpseudocode}
\usepackage{amsmath}
\usepackage{pifont}
\graphicspath{{media/}} 

\title{TableCenterNet: A one-stage network for table structure recognition
}

\author{ Anyi~Xiao, \\ School of Information Engineering \\ Nanchang Hangkong University\\ \texttt{2304081200003@stu.nchu.edu.cn}
\\
\And Cihui Yang\thanks{ \textbf{ Corresponding author.}} \\ School of Information Engineering \\ Nanchang Hangkong University\\ \texttt{yangcihui@nchu.edu.cn} \\
}

\begin{document}
\maketitle

\begin{abstract}
  Table structure recognition aims to parse tables in unstructured data into machine-understandable
  formats. Recent methods address this problem through a two-stage process or
  optimized one-stage approaches. However, these methods either require
  multiple networks to be serially trained and perform more time-consuming
  sequential decoding, or rely on complex post-processing algorithms to parse the
  logical structure of tables. They struggle to balance cross-scenario
  adaptability, robustness, and computational efficiency. In this paper, we
  propose a one-stage end-to-end table structure parsing network called
  TableCenterNet. This network unifies the prediction of table spatial and logical
  structure into a parallel regression task for the first time, and implicitly
  learns the spatial-logical location mapping laws of cells through a
  synergistic architecture of shared feature extraction layers and task-specific
  decoding. Compared with two-stage methods, our method is easier to train and
  faster to infer. Experiments on benchmark datasets show that TableCenterNet can
  effectively parse table structures in diverse scenarios and achieve state-of-the-art
  performance on the TableGraph-24k dataset. Code is available at
  \href{https://github.com/dreamy-xay/TableCenterNet}{https://github.com/dreamy-xay/TableCenterNet}.
\end{abstract}


\section{Introduction}
As the main carrier of structured data, tables are prevalent in diverse media such
as financial statements, academic papers, and natural-scene images, carrying
the organization and delivery of key information. With the deepening digital transformation,
the demand for accurately parsing the machine-understandable formats (including
the spatial and logical locations of cells) of tables from unstructured data
is constantly growing. Although humans can easily understand tables of various
styles and layouts, it remains challenging for automated systems.

\begin{figure*}[!htb]
  \centering
  \begin{subfigure}
    {\textwidth}
    \centering
    \includegraphics[width=0.82139\textwidth]{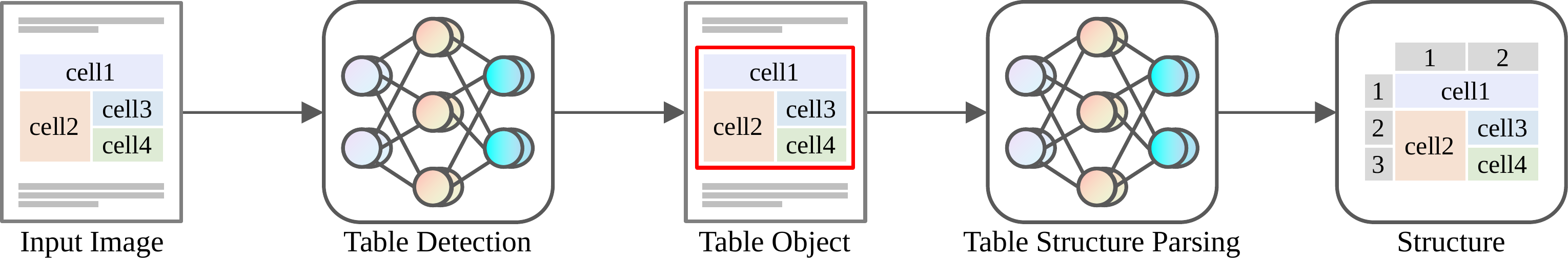}
    \caption{Previous two-stage approach based on object detection and table
      structure parsing.}
    \label{fig:MethodsCompareSub1}
    \vspace{0.5cm}
  \end{subfigure}
  \begin{subfigure}
    {\textwidth}
    \centering
    \includegraphics[width=\textwidth]{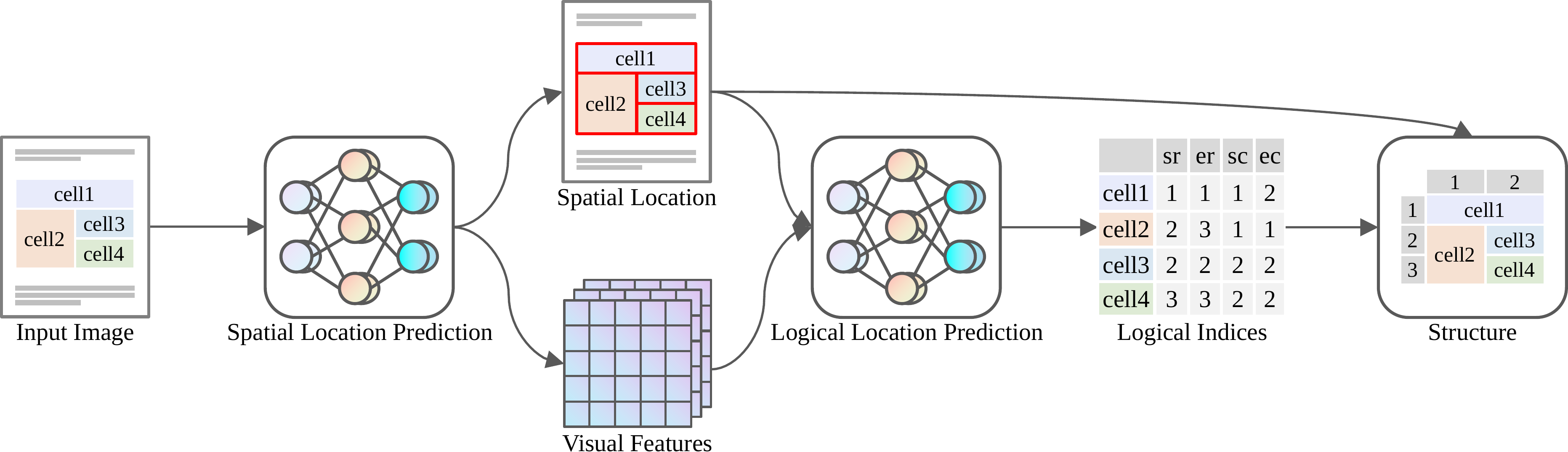}
    \caption{Previous two-stage approach based on spatial and logical location prediction. Here, $sr$, $er$, $sc$, $ec$ refer to the starting-row, ending-row, starting-column and ending-column respectively.}
    \label{fig:MethodsCompareSub2}
    \vspace{0.5cm}
  \end{subfigure}
  \begin{subfigure}
    {\textwidth}
    \centering
    \includegraphics[width=0.64495\textwidth]{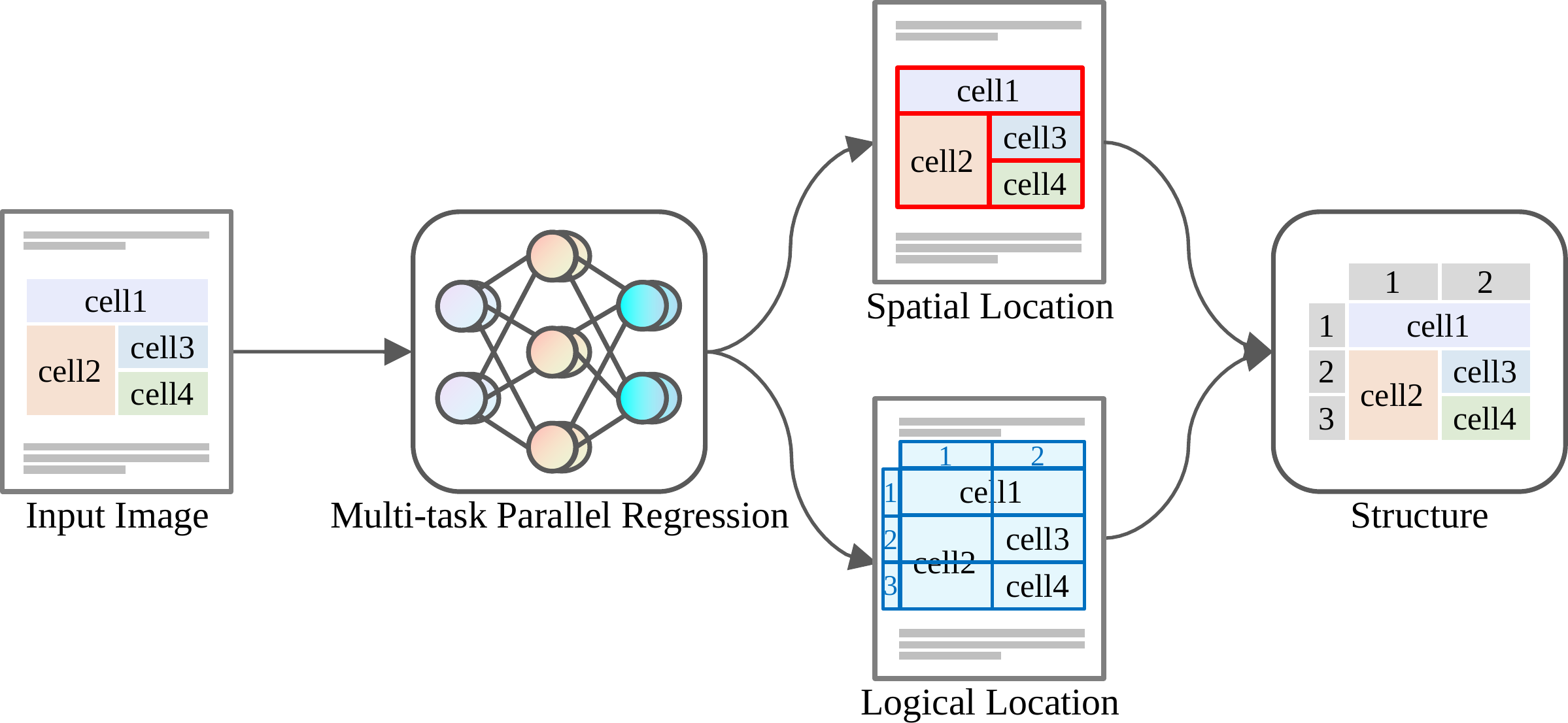}
    \caption{Proposed one-stage approach.}
    \label{fig:MethodsCompareSub3}
  \end{subfigure}
  \caption{Comparison of our one-stage approach with previous two-stage
    approaches.}
  \label{fig:MethodsCompare}
\end{figure*}

Previously, most table structure recognition (TSR) methods based on deep learning adopted
a two-stage processing flow. For example, studies such as
\cite{schreiber2017deepdesrt, zheng2021gte, Fernandes2024tablestrrec} first located the table region through object detection and then performed structural parsing on the cropped region, as illustrated
in \autoref{fig:MethodsCompareSub1}. Although these methods can handle some
standard document tables, they are difficult to cope with the problem of geometric
deformation in natural scenes due to shooting angle, occlusion or bending. In recent
years, some works have attempted to optimize the two-stage paradigm. For instance,
TGRNet \cite{xue2021tgrnet} and LORE \cite{xing2023lore} sequentially predicted
the spatial location and logical indices of cells through two networks, as shown
in \autoref{fig:MethodsCompareSub2}. While these methods perform well in
different scenarios, their phased design requires independent training of multiple
sub-modules to optimize performance, which significantly increases the
training complexity. Furthermore, the serial inference process introduces
additional computational overhead, making it difficult to meet the real-time requirements.
Although the existing one-stage methods simplify the process, they face bottlenecks
in scenario generalization. For example, TableNet \cite{paliwal2019tablenet}
rely on the assumption of a clean and structured background and cannot handle
blurring, uneven illumination, or geometric deformation in natural scenes. Although
Cycle-CenterNet \cite{long2021parsing} and SCAN \cite{wang2023scene} support table
parsing in natural scenes, it does not consider borderless tables and has
insufficient accuracy in complex scenarios such as cross-row and cross-column
merging. TRACE \cite{baek2023trace} is sensitive to geometric distortions such
as bending and relies on complex post-processing algorithms.  Overall, existing methods struggle to achieve an effective balance between cross-scenario adaptability, structural recognition robustness, and computational efficiency.

To address the above problems, we propose a one-stage end-to-end table structure
parsing network, TableCenterNet, based on the CenterNet \cite{xing2019centernet}
framework. As illustrated in \autoref{fig:MethodsCompareSub3}, this method unifies the spatial location detection and logical location
prediction of cells into a parallel regression task for the first time, and realizes efficient joint
learning through the synergistic architecture of a shared feature extraction layer
and task-specific decoding. This design not only avoids the error accumulation
problem of the two-stage approach, but also significantly improves the parsing
accuracy of complex tables (e.g., merging cells across rows and columns) by implicitly
learning the spatial-logical location mapping law. Experiments show that
TableCenterNet is highly competitive and outperforms existing state-of-the-art
methods for logical location prediction on the TableGraph-24k \cite{xue2021tgrnet}
dataset.

The main contributions of this paper are as follows:

\begin{itemize}
  \item We propose an end-to-end table structure parsing network that simultaneously
        predicts the spatial and logical locations of cells through parallel regression.

  \item Experiments on three public benchmark datasets containing tabular images
        of diverse scenarios demonstrate the effectiveness of our proposed method.

  \item We provide a simple and effective one-stage TSR method that simplifies
        the training process and accelerates inference, and the code is available
        to support further research on TSR.
\end{itemize}

\section{Related Work}
Early TSR methods \cite{laurentini1992identify, itonori1993table,
  thomas1998table, kieninger1999doc, shigarov2016conftable, rastan2019texus}
primarily rely on manually designed features and heuristic rules. For example,
they parse tables using visual cues such as text arrangement, lines, and
templates. These methods are effective only for tables in specific formats (e.g.,
PDF) and struggle to handle complex tables. With the rise of statistical
machine learning, approaches such as \cite{ng1999learning, wang2004table,
  liu2008identify, liu2009improving} attempt to reduce the reliance on rules,
but still require manual feature design, the strong assumptions regarding table
layouts restrict their practical applications. In recent years, deep learning-based
methods have shown better performance than traditional methods \cite{hashmi2021current}.
Based on the descriptions of these methods, we can broadly categorize them
into three groups: row and column division-based methods, markup sequence
generation-based methods, and cell reconstruction-based methods.

\subsection{Row and column division-based methods}
These methods mainly reconstruct the table structure by detecting or segmenting
row and column regions \cite{schreiber2017deepdesrt, paliwal2019tablenet,
  siddiqui2019deeptabstr, siddiqui2019rethinking, hashmi2021guided, qiao2021lgmpa,
  smock2022pubtabs-1m, Fernandes2024tablestrrec}, or by directly predicting the separation
lines for rows and columns \cite{tensmeyer2019deep, khan2019table,
  zou2020deepss, zhang2022split, lin2022tsrformer, ma2023robust, baek2023trace}.
For example, DeepTabStR \cite{siddiqui2019deeptabstr} improves the object detection
model using deformable convolutions to directly locate rows and columns;
DeepdeSRT \cite{schreiber2017deepdesrt} and TableNet \cite{paliwal2019tablenet}
utilize semantic segmentation models to predict row and column regions and
generate cells through intersection operations. However, none of the above
methods can handle the cases of row-spanning and column-spanning. Therefore,
SPLERGE \cite{tensmeyer2019deep} further proposes a two-stage strategy of segmentation
and merging, using a segmentation model to generate an initial cell grid and
then processing cross-cells through a merging module. But these methods rely on
the strict alignment assumption of rows and columns and have poor adaptability
to curved tables or non-axis-aligned layouts. Although RobusTabNet \cite{ma2023robust} enhances robustness to distorted tables by introducing a spatial CNN module \cite{pan2018spatial}, the heuristic rules in its post-processing stage may still affect the overall performance due to low-quality separator masks.

\subsection{Markup sequence generation-based methods}
Such methods \cite{deng2019challenges, li2020tablebank, zhong2020image, ye2021pingan,
  nassar2022tableformer} treat TSR as a generation task from
images to tag sequences and directly output structured tags such as HTML or LaTeX.
TableMaster \cite{ye2021pingan} is built on the Transformer architecture, which
extracts visual features through the encoder and generates HTML sequences via
the decoder. TableFormer \cite{nassar2022tableformer} not only decodes
structural tags but also decodes cell coordinates. However, these methods
usually rely on large-scale training data, and the length of the generated markup
sequence increases with greater table complexity, resulting in
performance degradation. In addition, since the model uses a sequential
decoding process, its inference is also more time-consuming.

\subsection{Cell reconstruction-based methods}
Cell reconstruction-based methods \cite{chi2019complicated, raja2020tsr,
  liu2021show, xue2021tgrnet, ichikawa2021image, long2021parsing, liu2022neural,
  xiao2022table, raja2022visual, wang2023scene, xing2023lore}
usually predicts table cells first and then utilize these cells as the basic
units to construct the global table structure by analyzing the associative
information between them. For example, GraphTSR \cite{chi2019complicated} models
the predicted cells as graph nodes and employs a graph attention network to predict
horizontal, vertical, or irrelevant relationships. FLAG-Net \cite{liu2021show} enhances
the reasoning ability for cell associations through both dense and sparse context
modeling. TGRNet \cite{xue2021tgrnet} and LORE \cite{xing2023lore} perform
logical location prediction on the detected cells using Graph Convolutional
Networks and Cascade Transformers, respectively, and directly generate logical
indices. Existing methods generally adopt a two-stage process of "detection-association".
Our method, however, makes full use of the cell location distribution
information in the feature map during the detection stage, performs parallel regression
of spatial and logical location, and directly matches the physical coordinates
with the logical locations through interpolation maps to obtain the logical
indices, thus simplifying the two-stage process into a one-stage end-to-end parsing.

\section{Methodology}

\subsection{Overall Architecture}
\begin{figure*}[!htb]
  \centering
  \includegraphics[width=1\textwidth]{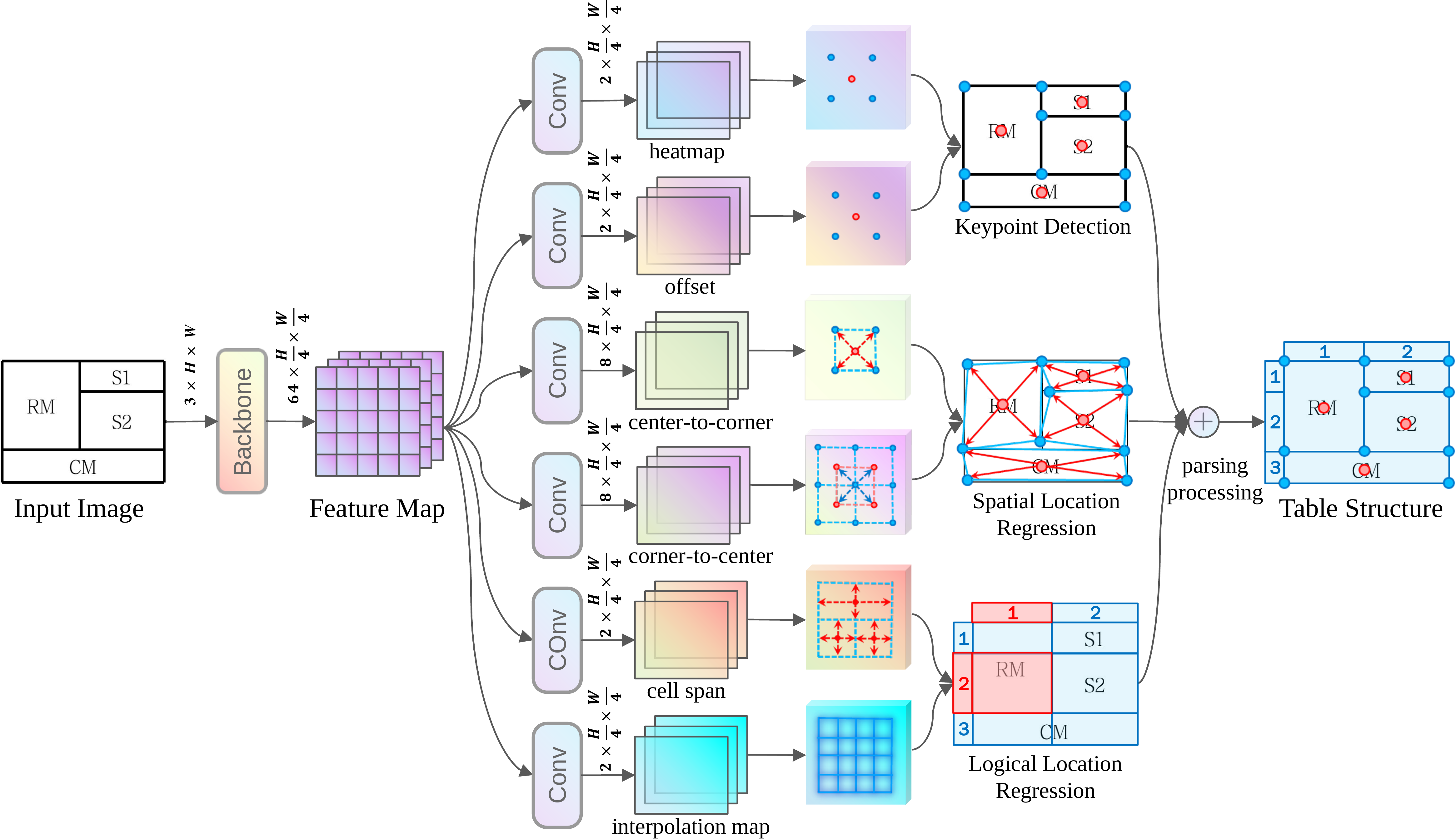}
  \caption{The architecture of our proposed method. Each "Conv" in the illustration
    represents a regression head, which comprises a $3\times3$ convolution
    followed by a $1\times1$ convolution.}
  \label{fig:TableCenterNet}
\end{figure*}

The overview of the proposed one-stage TableCenterNet is shown in \autoref{fig:TableCenterNet}. It employs a CNN backbone to extract visual features
of cells from the input image, and then jointly predicts the spatial locations
$\{\hat{b}_{1}, \hat{b}_{2}, ..., \hat{b}_{n}\}$ and logical locations
$\{\hat{l}_{1}, \hat{l}_{2}, ..., \hat{l}_{n}\}$ of cells in a multi-task
manner through six parallel regression heads. $n$ is the number of cells, and
the definitions of $\hat{b}_{i}$ and $\hat{l}_{i}$ are as follows:
\begin{equation}
  \hat{b}_{i}=\{(\hat{x}_{i,k},\hat{y}_{i,k})\}_{k=1,2,3,4}
\end{equation}
\vspace{-20pt}
\begin{equation}
  \hat{l}_{i}=\{\hat{r}^{st}_{i}, \hat{r}^{ed}_{i}, \hat{c}^{st}_{i}, \hat{c}^{ed}
  _{i}\}
\end{equation}
Where $(\hat{x}_{i,k},\hat{y}_{i,k})$ denotes the coordinates of the $k$-th corner
point of the $i$-th cell. The four corner points are arranged clockwise
starting from the upper-left corner. $\hat{r}^{st}_{i}$, $\hat{r}^{ed}_{i}$, $\hat
  {c}^{st}_{i}$, and $\hat{c}^{ed}$ correspond to the four logical indices of
the $i$-th cell: starting-row, ending-row, starting-column, and ending-column, respectively.

Specifically, TableCenterNet comprises three main components: key point detection,
spatial location regression, and logical location regression. The key point
detection aims to streamline the joint prediction of spatial and logical
location. Similar to CenterNet \cite{xing2019centernet}, we use two independent
heads to regress the offsets and heatmaps of key points. The two channels of
the key point heatmap correspond to the predictions of the cell center point
and the corner coordinates, respectively. For the heatmap
$\hat{Y}\in [0, 1]^{\frac{H}{4} \times \frac{W}{4}}$ of each channel, when $\hat
  {Y}_{x,y}= 1$, it indicates the position of a key point, while $\hat{Y}_{x,y}=0$
signifies the background area, where $H$ and $W$ denote the height and width
of the input image.

To regress the spatial locations of cells, we again incorporate two regression
heads: one to predict the vectors from the cell centers to the corner points, and
the other to predict the vectors from the corner points back to the centers.
The last two regression heads are used to predict the row and column span
information of the cells and to generate the logical location interpolation
map. All regression results will be parsed and processed sequentially, and
finally the table structure information will be output.

\subsection{Spatial Location Regression}
To enhance the accuracy of spatial location prediction, we adopted the Cycle-Pairing
Module \cite{long2021parsing}. By regressing the vectors
$\hat{u}_{i}=\{(\hat{x}_{i,k}^{u},\hat{y}_{i,k}^{u})\}_{k=1,2,3,4}$ from the center of each cell to its four corners (center-to-corners) and the vectors $\hat{v}_{i}=\{(\hat{x}_{i,k}^{v},\hat{y}_{i,k}^{v})\}_{k=1,2,3,4}$ from the four corners to the center (corners-to-center), and based on the center coordinates $\hat{t}_{i}=\{\hat{x}_{i}^{t},\hat{y}_{i}^{t}\}$ of each cell, we can calculate the approximate
spatial locations $\{\tilde{b}_{1}, \tilde{b}_{2}, ..., \tilde{b}_{n}\}$ of the
cells. The specific calculation formula is as follows:
\begin{equation}
  \tilde{b}_{i}    =\{(\tilde{x}_{i,k},\tilde{y}_{i,k})\}_{k=1,2,3,4},
  \begin{cases}
    \tilde{x}_{i,k}  = \hat{x}_{i,k}^{t}+ \hat{x}_{i,k}^{u} \\
    \tilde{y}_{i,k}  = \hat{y}_{i,k}^{t}+ \hat{y}_{i,k}^{u}
  \end{cases}
\end{equation}

In addition, we have optimized PairLoss \cite{long2021parsing} to work more effectively in conjunction with logical position regression, while reducing the interference caused by invalid corners-to-center vectors. The formula for the improved spatial location regression loss is as follows:
\begin{equation}
  \mathcal{L}_{spatial}=\frac{1}{8n}\sum_{i=1}^{n}{\omega(i)(\lambda_{u}\mathcal{L}_1(u_i,\hat{u}_i)+\lambda_{v}\mathcal{L}_1(v_i,\hat{v}_i))} + \lambda_e\mathcal{L}_{e}
\end{equation}
Among them, $u_{i}$ and $v_{i}$ represent the ground truth (GT) corresponding to the predicted
vectors $\hat{u}_{i}$ and $\hat{v}_{i}$, respectively. $\mathcal{L}_{1}$ denotes
the mean absolute error \cite{arnaud2016mae} loss function, while $\mathcal{L}_{e}$ represents the loss of invalid corners-to-center vectors. The hyperparameters $\lambda
  _{u}=1.0$, $\lambda_{v}=0.5$, and $\lambda_{e}=0.2$ are used to adjust the relative
importance of these loss items. Additionally, $\omega(i)$ dynamically weights the
$\mathcal{L}_{1}$ loss based on the quality of the regression, it is defined
as follows:
\begin{equation}
  \omega(i)=sin(\frac{\pi}{2}\cdot min(\frac{|u_{i}-\hat{u}_{i}|+|v_{i}-\hat{v}_{i}|}{|u_{i}|}
  ,1.0))
\end{equation}
Using the corners as reference points, each corner can point to the centers of up to four different cells, resulting in $4m$ corners-to-center vectors obtained through regression, where $m$ denotes the total number of cell corners.
Since $4m > 4n$, in addition to the vectors
$\{\hat{v}_{1}, \hat{v}_{2}, ..., \hat{v}_{n}\}$ matched by cells, there are also
many invalid corners-to-center vectors, which are defined as
$\hat{e}=\{(\hat{x}_{j}^{e}, \hat{y}_{j}^{e})\}_{j=1,2,...,4m-4n}$. From this, the loss $\mathcal{L}_{e}$ is formulated as follows:
\begin{equation}
  \mathcal{L}_{e}=\frac{1}{8m-8n}\sum_{j=1}^{4m-4n}{(|\hat{x}_j^e| + |\hat{y}_j^e|)}
\end{equation}

\subsection{Logical Location Regression}
Considering that the extracted feature maps already contain the location information
of cells, we capture the row-column demarcation features of the table by
regressing the corresponding logical location interpolation map, obtaining the
row interpolation map $\hat{I}_{r}\in \mathbb{R}^{\frac{H}{4} \times
  \frac{W}{4}}$ and the column interpolation map $\hat{I}_{c}\in \mathbb{R}^{\frac{H}{4}
  \times \frac{W}{4}}$. Meanwhile, based on the cell center, we regress the row-column
span $\hat{s}_{i}=\{\hat{s}_{i}^{r},\hat{s}_{i}^{c}\}$ of each cell, with the purpose
of synergistically supervise the regression of the interpolation map and simplify
the parsing process. Among them, $\hat{s}_{i}^{r}$ and $\hat{s}_{i}^{c}$ denote the row and column span of the $i$-th cell respectively.

\begin{figure*}[!htb]
  \centering
  \includegraphics[width=1\textwidth]{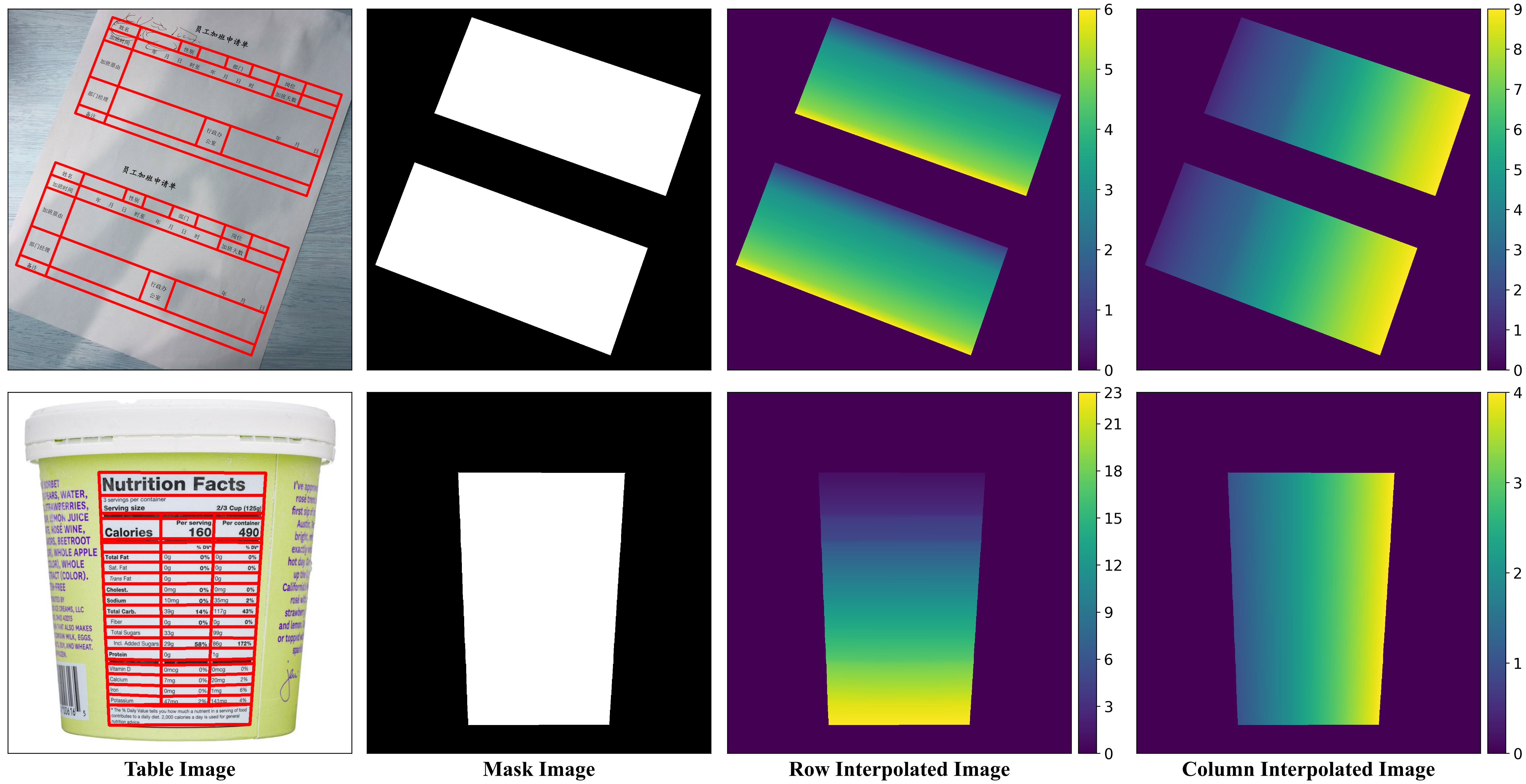}
  \caption{Visualization of row and column interpolation maps generated by the
    algorithm.}
  \label{fig:interpolatedImage}
\end{figure*}

\textbf{Generation of interpolation maps.} Before training, we generate corresponding
interpolation maps for each table image. For the GT of each image, we pair the
logical indices $l_{i}=\{r^{st}_{i}, r^{ed}_{i}, c^{st}_{i}, c^{ed}_{i}\}$ of each
cell with the physical coordinates $b_{i}=\{(x_{i,k},y_{i,k})\}_{k=1,2,3,4}$ to obtain
two sets of polygons $\mathcal{P}_{r}=\{P_{1}^{r},P_{2}^{r},... ,P_{n}^{r}\}$ and
$\mathcal{P}_{c}= \{P_{1}^{c}, P_{2}^{c},... ,P_{n}^{c}\}$ for row and column interpolation.
Where $P_{i}^{r}$ and $P_{i}^{c}$ can be defined as follows:
\begin{equation}
  P_{i}^{r}    = \{(x_{i,k},y_{i,k},o_{i,k}^{r})\}_{k=1,2,3,4},
  o_{i,k}^{r}  =
  \begin{cases}
    r^{st}_{i}    & \text{if }k = 1 \text{ or }k = 2 \\
    r^{ed}_{i}+ 1 & \text{otherwise}
  \end{cases}
\end{equation}
\begin{equation}
  P_{i}^{c}    = \{(x_{i,k},y_{i,k},o_{i,k}^{c})\}_{k=1,2,3,4},
  o_{i,k}^{c}  =
  \begin{cases}
    c^{st}_{i}    & \text{if }k = 1 \text{ or }k = 4 \\
    c^{ed}_{i}+ 1 & \text{otherwise}
  \end{cases}
\end{equation}
Then, we input $\mathcal{P}_{r}$ and $\mathcal{P}_{c}$ into \autoref{alg:interpolate} to calculate the corresponding row interpolation map
$I_{r}\in \mathbb{R}^{H \times W}$ and column interpolation map
$I_{c}\in \mathbb{R}^{H \times W}$. Since the physical coordinates of the input
cells are the same in the two calculations, the obtained mask image
$M \in \mathbb{R}^{H \times W}$ is also the same. The visualization results are
shown in \autoref{fig:interpolatedImage}. During training, we only need to reduce
$I_{r}$, $I_{c}$, and $M$ by a factor of 4 to use them as the GT for interpolation
map regression.

\begin{algorithm*}
  \caption{Polygons Interpolation}
  \label{alg:interpolate}
  \renewcommand{\algorithmicrequire}{\textbf{Input:}}
  \renewcommand{\algorithmicensure}{\textbf{Output:}}
  \begin{algorithmic}
    [1] \Require Polygons $\{P_{k}\}_{k=1}^n$ where $P_{k}= \{(x_{i}, y_{i}, o_{i}) | 1  \leq i \leq z \}$, Image dimensions $(H, W)$ \Ensure Interpolated image
    $I \in \mathbb{R}^{H \times W}$, Mask matrix $M \in \{0,1\}^{H \times W}$

    \State $I \gets \left[ \mathbf{0}\right]_{H \times W}$ \Comment{Initialize interpolated image}
    \State $M \gets \left[ \mathbf{0}\right]_{H \times W}$ \Comment{Initialize mask matrix}

    \For{$k = 1$ to $n$}
    \State $A_{k} \gets \frac{1}{2}\left|\sum_{i=1}^{z-1}(x_{i}y_{i+1}- x_{i+1} y_{i})\right|$ \Comment{Compute polygon area}
    \EndFor

    \State $\{P_{(k)}\}_{k=1}^n \gets \text{argsort}(\{A_{k}\}_{k=1}^n, \text{ascending})$ \Comment{Sort polygons by area}

    \For{each polygon $P_{(k)}$} \State $x_{\min}\gets \lfloor \min\{x_{i}\}_{i=1}^z \rfloor$,
    \quad $y_{\min}\gets \lfloor \min\{y_{i}\}_{i=1}^z \rfloor$
    \vspace{2pt}
    \State
    $x_{\max}\gets \lceil \max\{x_{i}\}_{i=1}^z \rceil$, \quad $y_{\max}\gets \lceil \max
      \{y_{i}\}_{i=1}^z \rceil$ \State $w \gets x_{\max}- x_{\min}+ 1$, \quad
    $h \gets y_{\max}- y_{\min}+ 1$
    \vspace{2pt}
    \State $\mathbf{X}\gets \{(x_{i}- x_{\min}, y_{i}- y_{\min}, i)\}_{i=1}^z$ \Comment{Collect and convert to local coordinates}
    \vspace{2pt}
    \State $\mathbf{O}\gets \{o_{i}\}_{i=1}^z$ \Comment{Collect vertex coordinate values}
    \vspace{2pt}
    \State $\mathbf{G}\gets \{(i,j) | 0 \leq i < w, 0 \leq j < h\}$ \Comment{Generate grid}

    \State $\mathbf{T}\gets \text{Triangulate}(\mathbf{X})$ \Comment{Delaunay triangulation}

    \State $\mathbf{Q}\gets \text{LinearInterpolation}(\mathbf{T}, \mathbf{O},
      \mathbf{G})$ \Comment{Default value is -1}

    \For{each $(i,j) \in \mathbf{G}$ with $q_{i,j}\in \mathbf{Q}$} \State
    $(i_{g}, j_{g}) \gets (i + x_{\min}, j + y_{\min})$ \Comment{Restore to global coordinates}
    \If{$q_{i,j}\geq 0$ \textbf{and} $M[j_{g}, i_{g}] = 0$} \State
    $I[j_{g}, i_{g}] \gets q_{i,j}$ \State $M[j_{g}, i_{g}] \gets 1$ \EndIf
    \EndFor \EndFor

    \State \Return $I$, $M$
  \end{algorithmic}
\end{algorithm*}

\textbf{Supervision of cell boundaries and spans.} When performing logical
location regression, two supervision methods are proposed to better understand
the correspondence between logical and spatial locations and the constraints between
logical locations. Supervision of cell boundary requires that all logical
boundaries of cells can effectively anchor the row-column demarcation of the
table and ensure the mutual exclusivity of the logical locations of different
cells. Specifically, the loss scheme for cell boundaries can be expressed as
follows:
\begin{equation}
  \mathcal{L}_{boundary}=\frac{\sum{f(I_{r})\mathcal{L}_{1}(I_{r}, \hat{I}_{r})}
    +\sum{f(I_{c})\mathcal{L}_{1} (I_{c},\hat{I}_{c})}}{2\sum{M}}
\end{equation}
Where $f(I_{r})$ and $f(I_{c})$ are weight functions used to enhance the supervision of the row-column demarcation of the table. In the interpolation map, the closer to the row-column demarcation, the higher the confidence of logical indices, and thus the weight should increase accordingly. $f(I_{r})$ and $f(I_{c})$ can be defined as follows:
\begin{equation}
  \begin{cases}
    f(I_{r})=(1.0-|I_{r}-round(I_{r})|)^{2} \\
    f(I_{c})=(1.0-|I_{c}-round(I_{c})|)^{2}
  \end{cases}
\end{equation}

Supervision of cell spans stipulates that the logical indices of each cell in the interpolation map should be consistent with its spans. Formally, the loss for constraining cell
spans can be defined as follows:
\begin{equation}
  \mathcal{L}_{span}=\frac{1}{2n}\sum_{i=1}^{n}{\mathcal{L}_s(i)}+ \frac{1}{4n}
  \sum_{i=1}^{n}{\tilde{\mathcal{L}}_s(i)}
\end{equation}
Among them, the loss $\mathcal{L}_{s}(i)$ of spanning regression and the loss
$\tilde{\mathcal{L}}_{s}(i)$ of spanning supervision for logical location of
the i-th cell can be described by the following formulas:
\begin{equation}
  \begin{cases}
    \mathcal{L}_{s}(i)=d_{r}(i)\mathcal{L}_{1}(s_{i}^{r}, \hat{s}_{i}^{r}) + d_{c}(i)\mathcal{L}_{1}(s_{i}^{c}, \hat{s}_{i}^{c}) \\
    \tilde{\mathcal{L}}_{s}(i)=d_{r}(i)\mathcal{L}_{1}(s_{i}^{r}, \tilde{s}_{i}^{r}) + d_{c}(i)\mathcal{L}_{1}(s_{i}^{c}, \tilde{s}_{i}^{c})
  \end{cases}
  \label{eq:spanLoss}
\end{equation}
In formula (\ref{eq:spanLoss}), $s_{i}=\{s_{i}^{r}, s_{i}^{c}\}$ denotes the
GT corresponding to $\hat{s}_{i}$, $d_{r}(i)$ and $d_{c}(i)$ are weight functions
used to calculate the average distance between $\hat{s}_{i}^{r}$ and $\tilde{s}
  _{i}^{r}$, which helps to coordinate the regression of the interpolation map
with the cell span and accelerates the convergence of the model. $d_{r}(i)$ and
$d_{c}(i)$ can be formulated as follows:
\begin{equation}
  \begin{cases}
    d_{r}(i)= sin(\frac{5\pi}{2}\cdot min(|\hat{s}_{i}^{r}-s_{i}^{r}|+|\tilde{s}_{i}^{r}-s_{i}^{r}|,0.2)) \\
    d_{c}(i)= sin(\frac{5\pi}{2}\cdot min(|\hat{s}_{i}^{c}-s_{i}^{c}|+|\tilde{s}_{i}^{c}-s_{i}^{c}|,0.2))
  \end{cases}
\end{equation}
Where the corresponding row span $\tilde{s}_{i}^{r}$ and column span
$\tilde{s}_{i}^{c}$ of the i-th cell in the interpolation maps $\hat{I}_{r}$
and $\hat{I}_{c}$ can be calculated by the following formulas:
\vspace{-5pt}
\begin{equation}
  \begin{cases}
    \tilde{s}_{i}^{r}= (\hat{I}_{r}[y_{i,4},x_{i,4}] - \hat{I}_{r}[y_{i,1},x_{i,1}], \hat{I}_{r}[y_{i,3},x_{i,3}] - \hat{I}_{r}[y_{i,2},x_{i,2}]) \\
    \tilde{s}_{i}^{c}=(\hat{I}_{c}[y_{i,2},x_{i,2}] - \hat{I}_{c}[y_{i,1},x_{i,1}], \hat{I}_{c}[y_{i,3},x_{i,3}] - \hat{I}_{c}[y_{i,4},x_{i,4}])
  \end{cases}
\end{equation}
\vspace{-5pt}

Then the regression loss of the logical location are as:
\begin{equation}
  \mathcal{L}_{logical}=\mathcal{L}_{boundary}+\mathcal{L}_{span}
\end{equation}

\subsection{Parsing Processing}
\begin{figure*}[!htb]
  \centering
  \begin{subfigure}
    {0.325\textwidth}
    \centering
    \includegraphics[width=\textwidth]{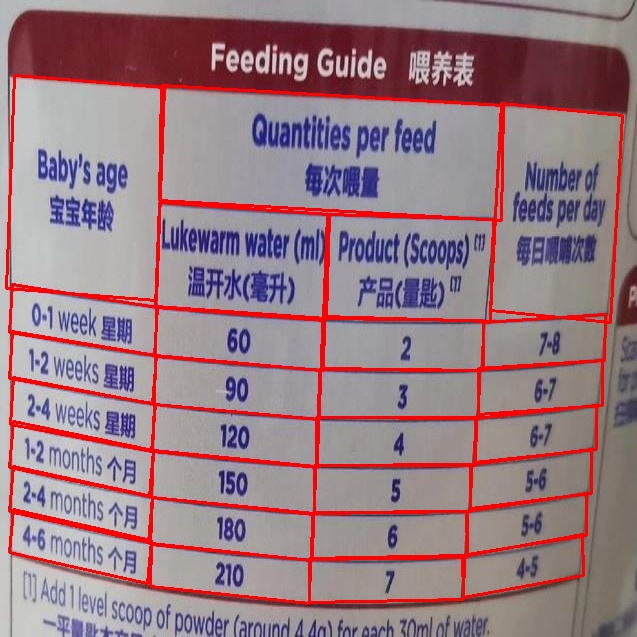}
    \caption{}
    \label{fig:SpaticalLocation1}
  \end{subfigure}
  \hfill
  \begin{subfigure}
    {0.325\textwidth}
    \centering
    \includegraphics[width=\textwidth]{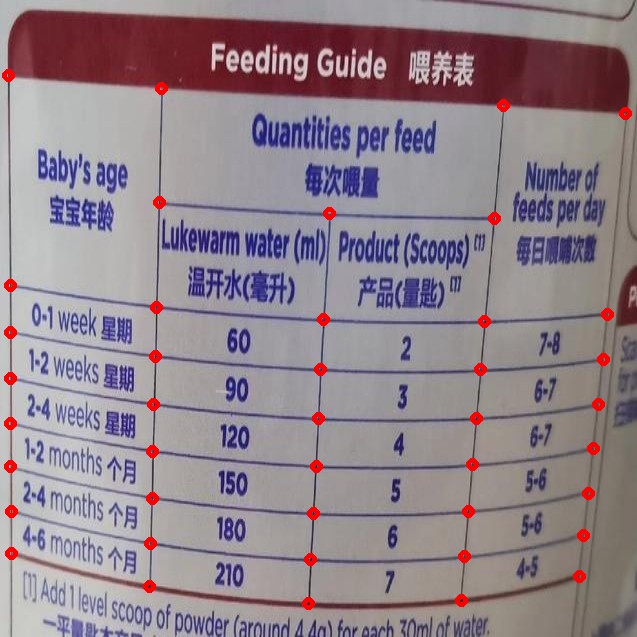}
    \caption{}
    \label{fig:SpaticalLocation2}
  \end{subfigure}
  \hfill
  \begin{subfigure}
    {0.325\textwidth}
    \centering
    \includegraphics[width=\textwidth]{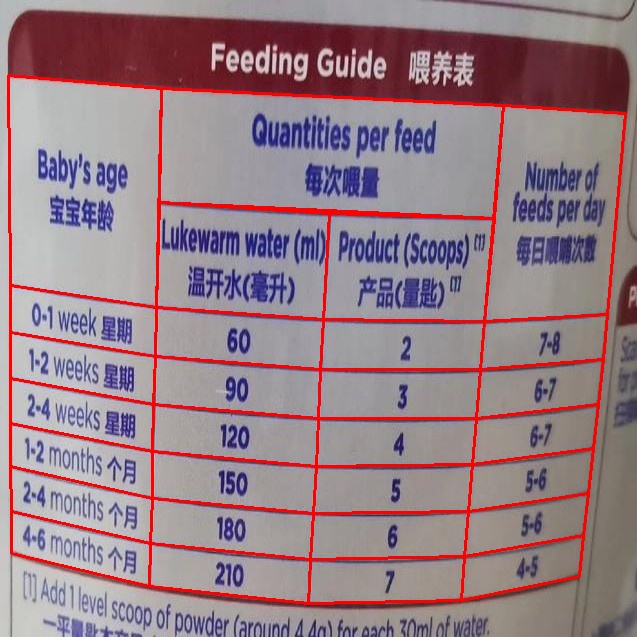}
    \caption{}
    \label{fig:SpaticalLocation3}
  \end{subfigure}
  \caption{Visualization of cell spatial location alignment. (a) is the result
    of cell spatial location regression, (b) is the result of cell corner point
    detection, and (c) is the result after cell spatial location alignment.}
  \label{fig:SpaticalLocation}
\end{figure*}

In \autoref{fig:SpaticalLocation1}, $\{\tilde{b}_{1}, \tilde{b}_{2}, ..., \tilde
  {b}_{n}\}$ obtained through regression are only approximations. The detected
cell corner points (see \autoref{fig:SpaticalLocation2}) need to be used
for alignment processing \cite{long2021parsing} to obtain more accurate spatial
locations $\{\check{b}_{1}, \check{b}_{2}, ..., \check{b}_{n}\}$, as shown in
\autoref{fig:SpaticalLocation3}. Then, calculate the physical coordinates
$\{\hat{b}_{1}, \hat{b}_{2}, ..., \hat{b}_{n}\}$ of all cells through 4x
magnification.

Subsequently, by directly matching the physical coordinates with the interpolation maps, the logical indices $\{\hat{l}_{1}, \hat{l}_{2}, ... , \hat{l}_{n}\}$ of all cells can be obtained. The matching process for the logical indices $\hat{l}_{i}=\{\hat{r}^{st}_{i}, \hat{r}^{ed}_{i}, \hat{c}^{st}_{i}, \hat{c}^{ed}_{i}\}$ of each cell can be formulated as follows:
\begin{equation}
  \begin{cases}
    \hat{r}^{st}_{i}= round(\hat{I}_{r}[\check{y}_{i,1}, \check{x}_{i,1}])                                      \\
    \hat{r}^{ed}_{i}= round(\hat{I}_{r}[\check{y}_{i,1}, \check{x}_{i,1}]) + \lfloor \hat{s}_{i}^{r}\rfloor - 1 \\
    \hat{c}^{st}_{i}= round(\hat{I}_{c}[\check{y}_{i,1}, \check{x}_{i,1}])                                      \\
    \hat{c}^{ed}_{i}= round(\hat{I}_{c}[\check{y}_{i,1}, \check{x}_{i,1}])+ \lfloor \hat{s}_{i}^{c}\rfloor - 1
  \end{cases}
\end{equation}
Where $(\check{y}_{i,1}, \check{x}_{i,1})$ denotes the coordinates of the upper-left
corner of the physical coordinates $\check{b}_{i}=\{(\check{y}_{i,k}, \check{x}_{i,k})\}
  _{k=1,2,3,4}$ of the i-th cell.

\subsection{Training Strategy}
\textbf{Design of overall loss function.} The proposed TableCenterNet is
trained in an end-to-end manner for three optimization tasks: keypoint detection,
spatial location regression and logical location regression. The global optimization
can be defined as:
\begin{equation}
  \mathcal{L}_{overall}= \mathcal{L}_{keypoint}+ \mathcal{L}_{spatical}+ \mathcal{L}
  _{logical}
\end{equation}
Where $\mathcal{L}_{keypoint}$ is the keypoint detection loss, which mainly consists
of the heat map loss $\mathcal{L}_{k}$ and offset loss $\mathcal{L}_{off}$ of keypoints,
consistent with CenterNet \cite{xing2019centernet}.

\begin{figure*}[!htb]
  \centering
  \includegraphics[width=0.9\textwidth]{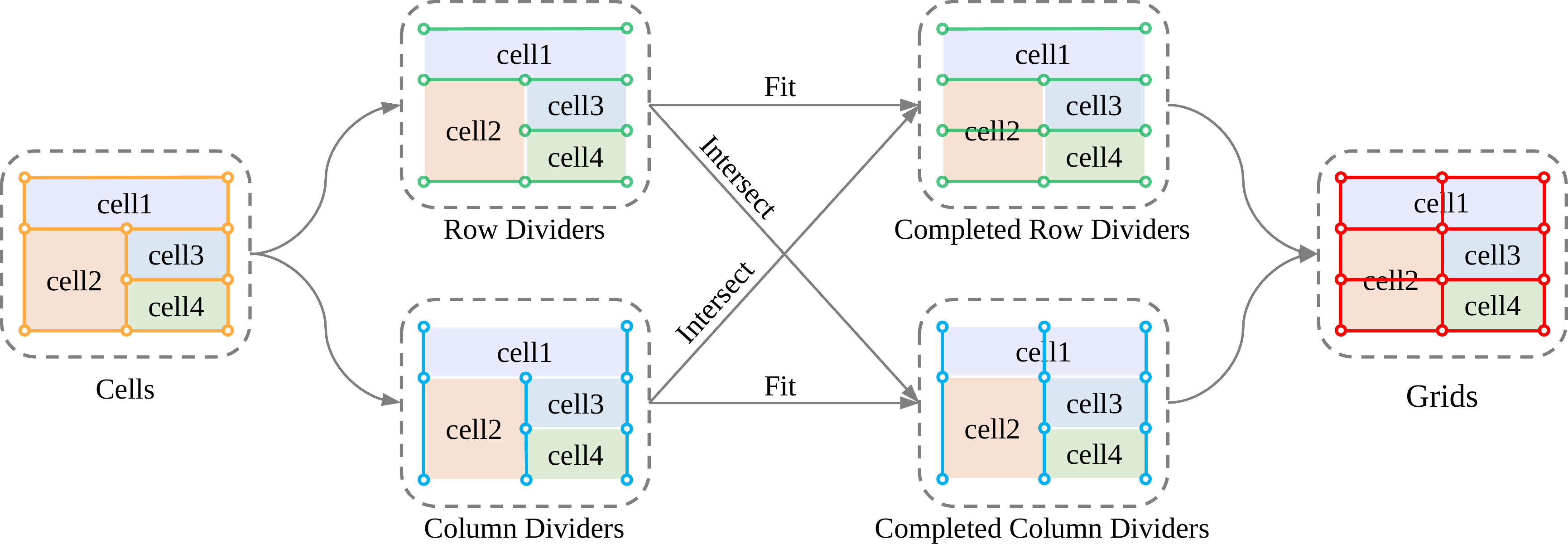}
  \caption{Flow of converting cells into grids. First, based on the physical coordinates and logical indexes of cell corners, row and column dividers are grouped. Then, the two types of dividers are completed by fitting and intersecting with each other. Finally, the completed row and column dividers are fused by corners to generate logical grids.}
  \label{fig:cellsToGrids}
\end{figure*}

\textbf{Selective gridding interpolation.} For datasets with unbiased prior
knowledge of physical and logical structures and no significant deformation, there
is no need to generate interpolation maps based on cells before training. As
shown in \autoref{fig:cellsToGrids}, the cells can be converted into a logical
grid structure, i.e., decomposing the cross-row and cross-column cells into
multiple logical grid cells. This effectively unifies the interpolation style of
each cell and improves the regression accuracy of the interpolation map. Instead of using cells to generate interpolation maps before training, grids are used.

\section{Experiments}

\subsection{Benchmark Datasets}

We conducted experiments on three popular public benchmarks, including the ICDAR2013 Table Competition dataset, the Wired Table in the Wild (WTW) dataset, and the TableGraph-24k dataset to validate the effectiveness of TableCenterNet.

\textbf{ICDAR-2013} \cite{gobel2013icdar} consists of 156 tables in PDF format from
EU/US government websites, which contain cross-cells and various other styles.
Because this paper focuses on reconstructing table from cells, an modified
version of ICDAR-2013, the ICDAR-2013.c \cite{smock2023aligning}, is used, which
provides detailed cell bounding boxes rather than word region boxes. It should
be noted that the original dataset does not specify the training and test sets,
we use 80\% of the table images for training and others for test following the
setting in
\cite{raja2020tsr, liu2022neural, liu2021show, xing2023lore, long2025lore}.

\textbf{WTW} \cite{long2021parsing} consists of digital document images, scanned
images, and images taken in the field, and contains a total of 10,970 training
images and 3,611 test images collected from wild complex scenes. This dataset
focuses on bordered tabular objects only and covers a wide range of
challenging situations such as inclined tables, curved tables, occluded and
blurred tables, extreme aspect ratio tables, overlaid tables, irregular tables,
and multicolored tables. The labeling of the dataset includes both the physical
and logical coordinates of the cells, with the physical coordinates
represented by quadrilaterals.

\textbf{TableGraph-24k} \cite{xue2021tgrnet} is a subset of TableGraph-350K, containing
20,000 training images, 2,000 validation images, and 2,000 test images. All
images in this dataset are sourced from scholarly articles on the arXiv
platform and include wired tables, wireless tables, and partial border tables.

\subsection{Evaluation Metrics}

Similar to previous studies \cite{raja2020tsr, xue2021tgrnet}, we first evaluate
the performance of our model in predicting the spatial locations of cells (physical
coordinates) using precision, recall, and F1 scores under the IoU threshold of
0.5. For the logical locations of cells (row/column information), we employed the
accuracy of logical location \cite{xue2019res2tim}, the F1 score of adjacency relationships
between cells \cite{gobel2012amethod, gobel2013icdar}, and Tree-Edit-Distance-based
Similarity (TEDS) \cite{zhong2020image} as evaluation metrics. The accuracy of logical
location and TEDS directly reflect the correctness of the predicted structure,
while the adjacency relationships only measure the quality of the intermediate
results of the structure \cite{xing2023lore}.

\subsection{Implementation Details}
All experiments were conducted in a PyTorch environment and executed on a workstation
equipped with two GeForce RTX 3090 GPUs. In these experiments, we respectively
adopted DLA-34 \cite{yu2018dla} and StartNet-s3 \cite{ma2024starnet} as the backbone
networks, and their weights were all initialized based on pre-trained models
for the ImageNet \cite{olga2015imagenet} classification task. Within the
CenterNet \cite{xing2019centernet} framework's regression network, we
configured the number of hidden channels for each regression head to $256$.
Model optimization was performed using the Adam algorithm with a batch size of
$22$ ($38$ for TableGraph-24k). For both backbone networks, the models were trained for $200$ epochs,
with the initial learning rate set to $1.25\times10^{-4}$. The learning rate was
decayed to $1.2 5\times10^{-5}$ and $1.25\times10^{-6}$ in the 140th and 180th
epochs, respectively. To ensure scale invariance, we uniformly resized the
input table images to a fixed size of $1024\times1024$ pixels, and $768\times76
  8$ pixels for the TableGraph-24k dataset, while maintaining the aspect ratio
of the images. After resizing, we also applied data augmentation and
normalization to the input images to accelerate the convergence of the network.

Due to the limited size of the ICDAR-2013 dataset, which contains only 258 table
images after preprocessing, we opted to perform $100$ rounds of fine-tuning on
the model trained on WTW using the segmented ICDAR-2013 training set. During the fine-tuning phase, considering that most tables in the WTW dataset have outer margins, we extend the original image by 100 black pixels on all sides before scaling the image. The initial learning rate decays to $1.25\times10^{-5}$ and
$1.25\times10^{-6}$ in the 70th and 90th epochs, respectively, while ensuring
that the other training configurations remain unchanged. In the testing phase,
we similarly expand the perimeter of the test image by $100$ black pixels
before performing inference.

\subsection{Results}
\begin{figure*}[!htb]
  \centering
  \includegraphics[width=\textwidth]{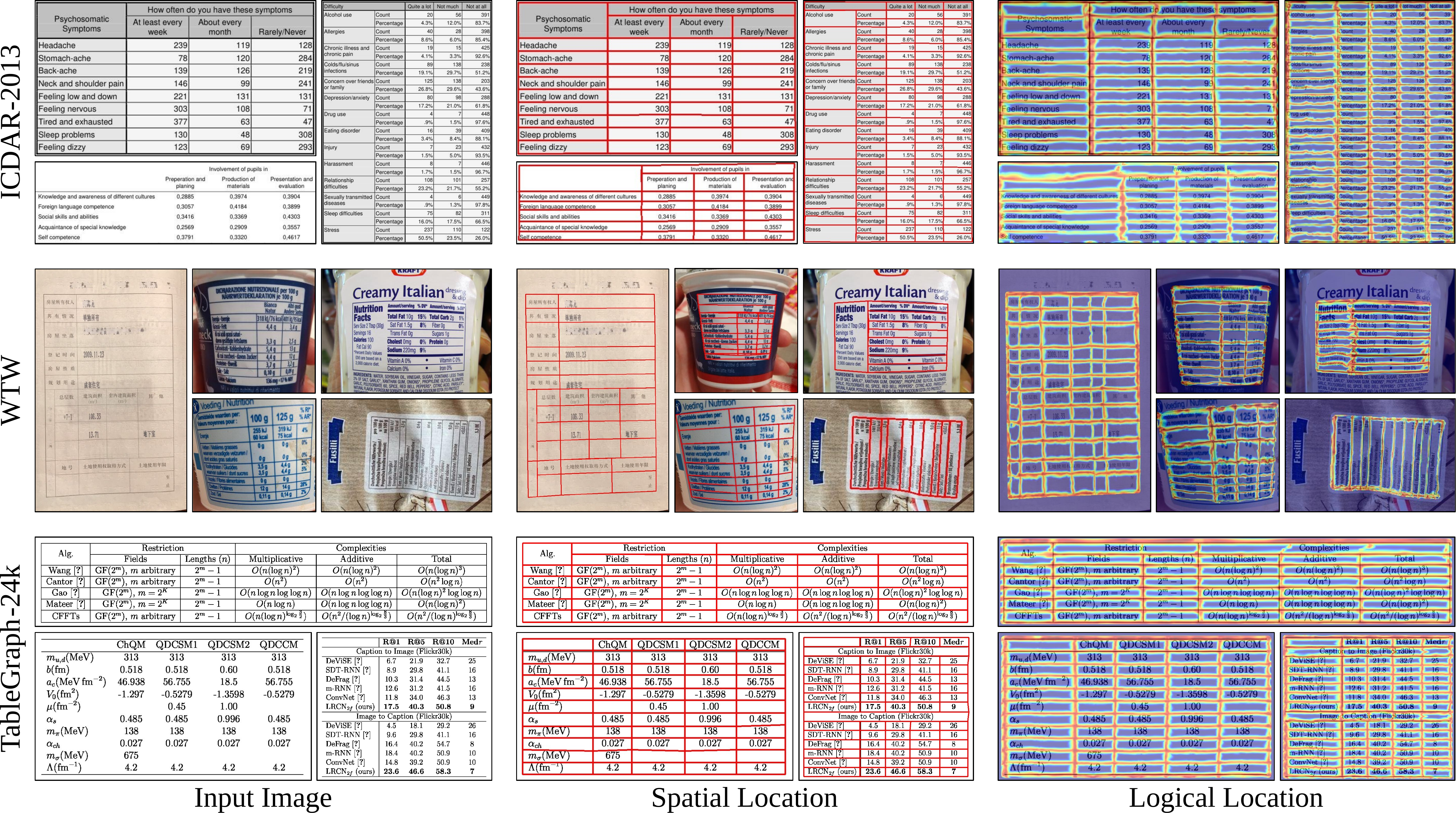}
  \caption{Qualitative results of our approach. The second column displays the
    spatial location of each cell. The third column visualizes the logical
    separation of rows and columns in the table using a heat map, which intensifies
    to red as the predicted interpolated map values approach positive integers.}
  \label{fig:quaResult}
\end{figure*}

To evaluate TableCenterNet's ability to recognize table structures in
different scenarios, we tested it on three benchmark datasets with different scenarios.
The qualitative results in \autoref{fig:quaResult} show that our method is
robust to complex structures, diverse colors, geometric transformations, and distorted
or deformed tables in both photographic and scanning scenes. Furthermore, our method
also demonstrates generalization to digital native scenarios, effectively
handling tables across cells, with bounded cells, and without bounded cells.

\textbf{Results on ICDAR-2013.} As shown in \autoref{tab:Icdar2013Result},
our proposed TableCenterNet method outperforms most previous methods in
recognizing both the physical and logical structures of tables. Compared to the
robust baseline model LORE \cite{xing2023lore}, TableCenterNet-D enhances the physical
coordinate F1 scores and the accuracy of the logical location by 0.6\% and 4.3\%,
respectively. Similarly, TableCenterNet-S improves these two metrics by 0.3\% and
1.3\%, respectively. For comparison with Cycle-CenterNet \cite{long2021parsing},
we used a model trained on the WTW dataset to test the wired tables in ICDAR-2013 \cite{smock2023aligning}.
The results indicate that our method exceeds Cycle-CenterNet \cite{long2021parsing}
and SCAN \cite{wang2023scene} in terms of precision, recall, and F1 score
metrics for adjacency. This shows that our method achieves competitive results
in both wired and wireless scenarios.

\begin{table*}[!htb]
  \caption{Comparison of the accuracy of logical location on ICDAR-2013. The
    precision, recall and F1 score are evaluated on physical coordinates-based and
    adjacency relation-based metrics. Here, $\dagger$ denotes for the result on wired tables
    only, whereas $\ddagger$ means that pre-trained data are used. Bolds denote the best. Underlines denote the best under $\dagger$ condition.}
  \centering
  \begin{tabular}{lccccccccc}
    \toprule \multirow{2}{*}{Model}                 & \multirow{2}{*}{Backbone} & \multirow{2}{*}{Training Datasets} & \multicolumn{3}{c}{Physical Coordinates} & \multicolumn{3}{c}{Adjacency Relation} & \multirow{2}{*}{Acc}                                                                          \\
    \cmidrule(lr){4-6} \cmidrule(lr){7-9}           &                           &                                    & P                                        & R                                      & F1                   & P                & R                & F1               &               \\
    \midrule ReS2TIM \cite{xue2019res2tim}          & -                         & ICDAR2013$\ddagger$                & -                                        & -                                      & -                    & 73.4             & 74.7             & 74.0             & 17.4          \\
    TGRNet \cite{xue2021tgrnet}                     & ResNet-50                 & ICDAR2013$\ddagger$                & 68.2                                     & 65.2                                   & 66.7                 & -                & -                & -                & 27.5          \\
    Cycle-CenterNet \cite{long2021parsing}          & DLA-34                    & WTW+ICDAR2019                      & -                                        & -                                      & -                    & 95.5             & 88.3             & 91.7             & -             \\
    TabStrNet \cite{raja2020tsr}                    & ResNet-101                & SciTSR                             & -                                        & -                                      & -                    & 93.0             & 90.8             & 91.9             & -             \\
    LGPMA \cite{qiao2021lgmpa}                      & ResNet-50                 & ICDAR2013$\ddagger$                & -                                        & -                                      & -                    & 96.7             & 99.1             & 97.9             & -             \\
    Cycle-CenterNet$\dagger$ \cite{long2021parsing} & DLA-34                    & WTW                                & -                                        & -                                      & -                    & 97.5             & 98.4             & 98.0             & -             \\
    TOD \cite{raja2022visual}                       & -                         & FinTabNet                          & -                                        & -                                      & -                    & 98.0             & 97.0             & 98.0             & -             \\
    SCAN$\dagger$ \cite{wang2023scene}              & ResNet-50                 & WTW                                & -                                        & -                                      & -                    & 98.2             & 98.1             & 98.2             & -             \\
    FLAGNet \cite{liu2021show}                      & ResNet-50                 & ICDAR2013$\ddagger$                & -                                        & -                                      & -                    & 97.9             & 99.3             & 98.6             & -             \\
    NCGM \cite{liu2022neural}                       & ResNet-18                 & ICDAR2013$\ddagger$                & -                                        & -                                      & -                    & 98.4             & 99.3             & 98.8             & -             \\
    LORE \cite{xing2023lore}                        & DLA-34                    & ICDAR2013$\ddagger$                & -                                        & -                                      & 97.2                 & 99.2             & 98.6             & 98.9             & 86.8          \\
    \midrule TableCenterNet-D                       & DLA-34                    & ICDAR2013$\ddagger$                & \textbf{98.7}                            & \textbf{96.9}                          & \textbf{97.8}        & 96.1             & 99.1             & 97.6             & \textbf{91.1} \\
    TableCenterNet-D$\dagger$                       & DLA-34                    & WTW                                & 97.1                                     & 97.9                                   & 97.5                 & \underline{98.4} & \underline{98.6} & \underline{98.5} & 94.6          \\
    TableCenterNet-S                                & StartNet-s3               & ICDAR2013$\ddagger$                & 98.2                                     & 96.8                                   & 97.5                 & 93.7             & 97.5             & 95.6             & 88.1          \\
    TableCenterNet-S$\dagger$                       & StartNet-s3               & WTW                                & 97.1                                     & 98.2                                   & 97.6                 & \underline{98.4} & \underline{98.6} & \underline{98.5} & 94.1          \\
    \bottomrule
  \end{tabular}
  \label{tab:Icdar2013Result}
\end{table*}

\textbf{Results on WTW.} Considering that the previous methods used mainly two
IoU thresholds, 0.5 and 0.9, to evaluate the prediction performance of
physical coordinates, we conducted independent evaluations under these two distinct
thresholds. The experimental results are presented in \autoref{tab:WTWResult}. The physical coordinate F1 scores of TableCenterNet-D and
TableCenterNet-S are both improved by 0.9\% compared to the optimal LORE \cite{xing2023lore}
at an IoU threshold of 0.5, and the physical coordinate F1 scores at an IoU
threshold of 0.9 demonstrated improvements of 13\% and 12.6\% over Cycle-CenterNet \cite{long2021parsing},
respectively. For the evaluation of cell logical coordinates, TableCenterNet-D
outperforms LORE \cite{xing2023lore} in both adjacency and the accuracy of logical
location. In addition, TableCenterNet-D and TableCenterNet-S improve by 1\%
and 0.4\% in TEDS compared to the optimal method \cite{wang2023scene},
indicating that our approach also offers parsing advantages for complex tables
in the wild.

\begin{table*}[!htb]
  \caption{Comparison of the accuracy of logical location and TEDS on WTW. The
    precision, recall and F1 score are evaluated on physical coordinates-based
    and adjacency relation-based metrics. Here $\dagger$ means using an IoU threshold of
    0.9 to evaluate the performance of physical coordinates. Bolds denote the best according to the 'Evaluation Metrics' section.
    Underlines denote the best under $\dagger$ condition.}
  \centering
  \begin{tabular}{lccccccccc}
    \toprule \multirow{2}{*}{Model}                       & \multirow{2}{*}{Backbone} & \multicolumn{3}{c}{Physical Coordinates} & \multicolumn{3}{c}{Adjacency Relation} & \multirow{2}{*}{Acc} & \multirow{2}{*}{TEDS}                                                                 \\
    \cmidrule(lr){3-5} \cmidrule(lr){6-8}                 &                           & P                                        & R                                      & F1                   & P                     & R             & F1            &               &               \\
    \midrule CascadeTabNet \cite{prasad2020cascadetabnet} & CascadeNet                & -                                        & -                                      & -                    & 16.4                  & 3.6           & 5.9           & -             & 11.4          \\
    Split+Heuristic$\dagger$ \cite{tensmeyer2019deep}     & -                         & 3.2                                      & 3.6                                    & 3.4                  & 25.7                  & 29.9          & 27.6          & -             & 26.0          \\
    TGRNet \cite{xue2021tgrnet}                           & ResNet-50                 & -                                        & -                                      & 64.7                 & -                     & -             & -             & 24.3          & -             \\
    FLAGNet \cite{liu2021show}                            & -                         & -                                        & -                                      & -                    & 91.6                  & 89.5          & 90.5          & -             & -             \\
    Cycle-CenterNet$\dagger$ \cite{long2021parsing}       & DLA-34                    & 78.0                                     & 78.5                                   & 78.3                 & 93.3                  & 91.5          & 92.4          & -             & 83.3          \\
    TSRFormer \cite{lin2022tsrformer}                     & ResNet-18                 & -                                        & -                                      & -                    & 93.7                  & 93.2          & 93.4          & -             & -             \\
    NCGM \cite{liu2022neural}                             & ResNet-18                 & -                                        & -                                      & -                    & 93.7                  & 94.6          & 94.1          & -             & -             \\
    SCAN \cite{wang2023scene}                             & ResNet-50                 & -                                        & -                                      & -                    & -                     & -             & -             & -             & 90.7          \\
    TRACE$\dagger$ \cite{baek2023trace}                   & ResNet-50                 & 63.8                                     & 65.7                                   & 64.8                 & 93.5                  & 95.5          & 94.5          & -             & -             \\
    LORE \cite{xing2023lore}                              & DLA-34                    & -                                        & -                                      & 96.4                 & 94.5                  & 95.9          & 95.1          & 82.9          & -             \\
    \midrule TableCenterNet-D                             & DLA-34                    & \textbf{98.1}                            & \textbf{96.5}                          & \textbf{97.3}        & \textbf{94.8}         & \textbf{97.2} & \textbf{96.0} & \textbf{83.0} & \textbf{91.7} \\
    TableCenterNet-D$\dagger$                             & DLA-34                    & \underline{91.9}                         & \underline{90.8}                       & \underline{91.3}     & -                     & -             & -             & -             & -             \\
    TableCenterNet-S                                      & StartNet-s3               & \textbf{98.1}                            & \textbf{96.5}                          & \textbf{97.3}        & 94.3                  & 96.9          & 95.6          & 81.5          & 91.1          \\
    TableCenterNet-S$\dagger$                             & StartNet-s3               & 91.4                                     & 90.4                                   & 90.9                 & -                     & -             & -             & -             & -             \\
    \bottomrule
  \end{tabular}
  \label{tab:WTWResult}
\end{table*}

\textbf{Results on TableGraph-24k.} To our knowledge, only two methods have
been trained and tested on this data set for the precision of cell physical
and logical coordinates: TGRNet \cite{xue2021tgrnet} and LORE \cite{xing2023lore},
respectively. In \autoref{tab:TG24kResult}, we observe that TableCenterNet
significantly outperforms previous methods. When the F1 scores for the physical
coordinates of the cells are quite similar compared to the optimal method \cite{xing2023lore},
TableCenterNet-D and TableCenterNet-S have improved the accuracy of logical
location by 7.2\% and 6.4\%, respectively, achieving state-of-the-art
performance. This demonstrates the substantial advantage of our methods in the
digital native table image recognition scenario for scientific articles.

\begin{table*}[!htb]
  \caption{Comparison of the accuracy of logical location on TableGraph-24k. Acc$_{rowSt}$, Acc$_{rowEd}$, Acc$_{colSt}$, and Acc$_{colEd}$ refer to the accuracy of the four logical indices of the starting-row, starting-column, ending-row, and ending-column, respectively. The
    precision, recall and F1 score are evaluated on physical coordinates-based
    metrics. Bolds denote the best.}
  \centering
  \begin{tabular}{lccccccccc}
    \toprule \multirow{2}{*}{Model}          & \multirow{2}{*}{Backbone} & \multicolumn{3}{c}{Physical Coordinates} & \multicolumn{5}{c}{Logical Location}                                                                                                 \\
    \cmidrule(lr){3-5} \cmidrule(lr){6-10} ~ & ~                         & P                                        & R                                    & F1            & Acc$_{rowSt}$ & Acc$_{rowEd}$ & Acc$_{colSt}$ & Acc$_{colEd}$ & Acc           \\
    \midrule TGRNet \cite{xue2021tgrnet}     & ResNet-50                 & 91.6                                     & 89.5                                 & 90.6          & 91.7          & 91.6          & 91.9          & 92.3          & 83.2          \\
    LORE \cite{xing2023lore}                 & DLA-34                    & -                                        & -                                    & 96.1          & -             & -             & -             & -             & 87.9          \\
    \midrule TableCenterNet-D                & DLA-34                    & 95.4                                     & \textbf{96.6}                        & 96.0          & \textbf{97.9} & \textbf{97.9} & \textbf{97.3} & \textbf{97.1} & \textbf{95.1} \\
    TableCenterNet-S                         & StarNet-s3                & \textbf{95.7}                            & \textbf{96.6}                        & \textbf{96.2} & 97.8          & 97.7          & 96.7          & 96.4          & 94.3          \\
    \bottomrule
  \end{tabular}
  \label{tab:TG24kResult}
\end{table*}

\subsection{Ablation Study}
In this subsection, we isolate the contributions of the key components in
TableCenterNet and conduct ablation experiments on the WTW dataset. In all the
experiments presented in \autoref{tab:ablation}, the differences in spatial
location accuracy are minimal, so we focus on the effect of key components on logical
location accuracy.

\begin{figure*}[!htb]
  \centering
  \includegraphics[width=1\textwidth]{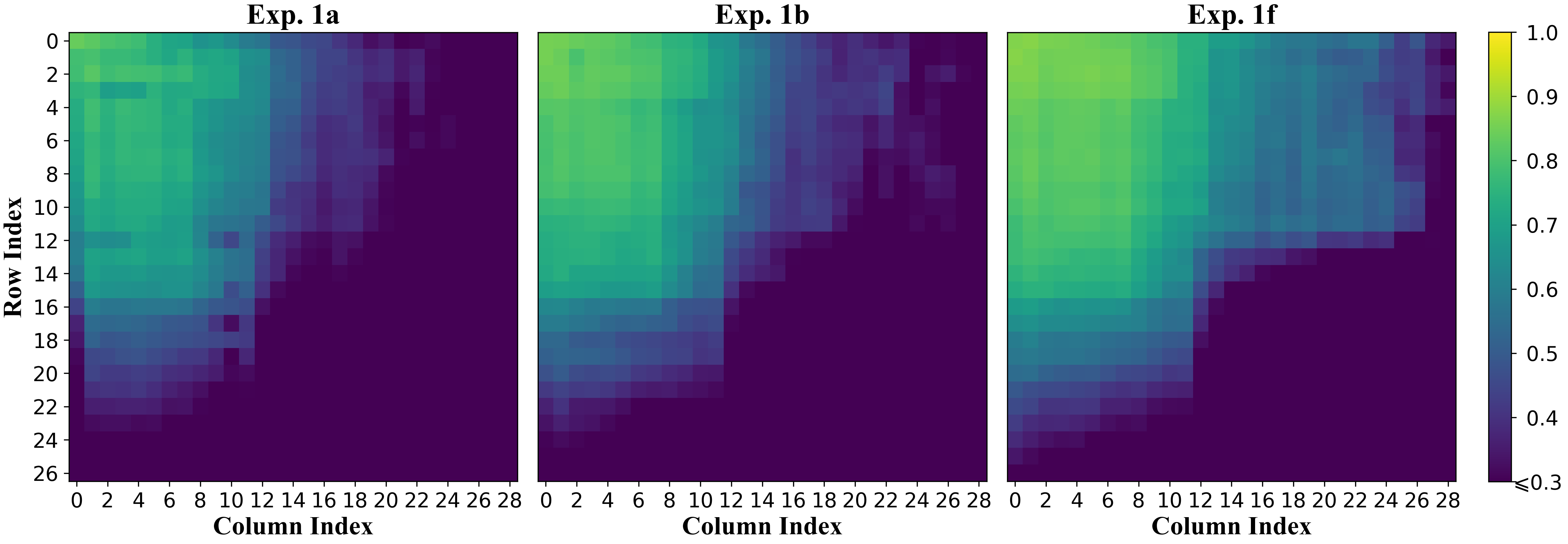}
  \caption{Visualization of the accuracy on each logical location in ablation
    study. Only the parts with a relatively large proportion in the logical indices
    are displayed.}
  \label{fig:ablation}
\end{figure*}

\textbf{Effectiveness of cell boundaries and spans supervisions.} Based on the
results of Exp.1a and Exp.1b, we observe that the cell-span regression yields a
performance improvement of +4.4\% Acc. Comparing Exp.1a and Exp.1c, the cell span
supervised loss $L_{span}$ leads to an even more substantial enhancement, with
a performance improvement of +6.5\% Acc. This is due to the fact that $L_{span}$
provides anchors for interpolation plot regression and strengthens the constraints
on the range of interpolation between cells. From Exp.1d and Exp.1f, the cell boundary
supervised loss $L_{boundary}$ also improves by +0.6\%Acc due to the fact that
it reduces the interpolation interference. According to the properties of interpolation
map generation, the logical indices of cell boundaries are closer to the true
values, while the logical indices of in-cell interpolation are approximate and
less reliable. In \autoref{fig:ablation}, we visualize the accuracy of each
logical location by heat map. From the three heat maps, we can find that cell spanning
regression improves the accuracy of small logical indices. When the two
supervision mechanisms are combined, the accuracy of large logical indices is also
significantly improved.

\textbf{Impact of different logical location regression methods.} In Exp.2a and Exp.2b, we replace interpolation map regression with logical indices regression of the upper-left corner point and the four corner points respectively. Both of them use the optimal loss computation scheme, and the network structure remains consistent with TableCenterNet. Compared with Exp.2f, the performance of these
two logical location regression methods decreases by 3.2\%Acc and 1.2\%Acc, respectively,
indicating that the interpolation map regression can better model the
dependencies and constraints between logical locations of different cells. When
there is an offset in the spatial locations regressed by the methods in Exp.2a and Exp.2b, the logical locations are prone to errors, whereas interpolation
maps can ensure the accuracy of logical indices as much as possible by rounding
the interpolation results.

\begin{table*}[!htb]
  \caption{Ablation study of TableCenterNet. The precision, recall and F1
    score are evaluated on physical coordinates-based metrics. A-c, A-r and Acc
    refer to the accuracy of column indices, row indices and all logical indices.
    All these models are trained from scratch according to the 'Implementation
    Details' section.}
  \centering
  \begin{tabular}{lcccccccccc}
    \toprule \multirow{2}{*}{Exp.}                                               & \multicolumn{2}{c}{Regression} & \multicolumn{2}{c}{Supervisions} & \multicolumn{3}{c}{Spatial Location} & \multicolumn{3}{c}{Logical Location}                                           \\
    \cmidrule(lr){2-3} \cmidrule(lr){4-5} \cmidrule(lr){6-8} \cmidrule(lr){9-11} & Logical Indices                & Cell Span                        & $L_{span}$                           & $L_{boundary}$                       & P    & R    & F1   & A-c  & A-r  & Acc  \\
    \midrule 1a                                                                  & Interpolation map              & \ding{55}                        & \ding{55}                            & \ding{55}                            & 98.1 & 96.4 & 97.3 & 86.4 & 84.8 & 75.1 \\
    1b                                                                           & Interpolation map              & \ding{51}                        & \ding{55}                            & \ding{55}                            & 98.1 & 96.4 & 97.2 & 89.0 & 87.4 & 79.5 \\
    1c                                                                           & Interpolation map              & \ding{55}                        & \ding{51}                            & \ding{55}                            & 98.2 & 96.5 & 97.4 & 90.7 & 88.3 & 81.6 \\
    1d                                                                           & Interpolation map              & \ding{51}                        & \ding{51}                            & \ding{55}                            & 98.1 & 96.5 & 97.3 & 91.0 & 88.8 & 82.4 \\
    1f                                                                           & Interpolation map              & \ding{51}                        & \ding{51}                            & \ding{51}                            & 98.1 & 96.5 & 97.3 & 91.4 & 89.1 & 83.0 \\
    \midrule 2a                                                                  & Upper-left corner point        & \ding{51}                        & -                                    & -                                    & 97.8 & 96.4 & 97.1 & 89.6 & 87.0 & 79.8 \\
    2b                                                                           & Four corner points             & \ding{51}                        & \ding{51}                            & -                                    & 97.8 & 96.5 & 97.1 & 90.9 & 88.2 & 81.8 \\
    \bottomrule
  \end{tabular}
  \label{tab:ablation}
\end{table*}

\subsection{Computational Analysis}
We compare three measures of the computational efficiency of the models in \autoref{tab:ComputationalAnalysis}, provided that the comprehensive performance is comparable. The input image size is $1024 \times 1024$, and both the comprehensive performance and the average inference time are computed based on the WTW \cite{long2021parsing} test set, where the comprehensive performance is referenced from \cite{xue2021tgrnet}, based on a combination of the F1 of the physical coordinates and the accuracy of logical locations. With the same backbone, TableCenterNet-D reduces the number of parameters by 27.3\% compared to LORE, but increases the inference speed by 15.7 times. TableCenterNet-S further improves the computational efficiency by using the StarNet-s3 \cite{ma2024starnet} backbone. Compared with TableCenterNet-D, the number of model parameters and FLOPs of TableCenterNet-S are reduced by 62.7\% and 62.2\%, respectively, which makes it possible to deploy on edge devices.


\begin{table*}[!htb]
  \caption{Comparison between TableCenterNet and LORE in terms of the number of parameters, FLOPs, and average inference time when the comprehensive performance F$_{\beta=0.5}$ is comparable. The units are million for the
    number of parameters, giga for the FLOPs, and milliseconds for the average inference
    time.}
  \centering
  \begin{tabular}{lccccc}
    \toprule Model   & Backbone   & F$_{\beta=0.5}$$\uparrow$ & Parameters$\downarrow$ & FLOPs$\downarrow$ & Average Inference Time$\downarrow$ \\
    \midrule LORE    & DLA-34     & 0.934                     & 27.86                  & 142.3             & 3788.4                             \\
    TableCenterNet-D & DLA-34     & 0.941                     & 16.82                  & 133.1             & 227.4                              \\
    TableCenterNet-S & StarNet-s3 & 0.937                     & 6.27                   & 50.3              & 214.8                              \\
    \bottomrule
  \end{tabular}
  \label{tab:ComputationalAnalysis}
\end{table*}

\section{Conclusion}
In this paper, we present a one-stage end-to-end table structure parsing
method, TableCenterNet, which uses a single model to simultaneously regress
both the spatial and logical structures of tables. Unlike previous methods \cite{xue2021tgrnet,
  xing2023lore}
that focus on logical location regression, TableCenterNet directly regresses the
distribution of the logical structure of each table at various locations within
the input image only through convolution along with supervision of cell boundaries
and spans, simplifying the model training process and accelerating model
inference. Experimental results demonstrate that our approach achieves competitive
outcomes across diverse scenarios, including the challenging table scenario, and
achieves state-of-the-art performance on the TableGraph-24k dataset. In future
work, we plan to collect challenging wireless tabular data from diverse scenarios
to further train our models and provide a reference for improvement.

\section*{Acknowledgments}
This work is supported by the Natural Science Foundation of China (No. 62366034).

\bibliographystyle{unsrt}
\bibliography{references}
\end{document}